\documentclass[letterpaper, 10 pt, conference]{ieeeconf}  % Comment this line out if you need a4paper

\usepackage{epsfig}
\usepackage{graphicx}
\usepackage{amsmath}
\usepackage{amssymb}
\usepackage{subfigure}
\usepackage{multirow}

\IEEEoverridecommandlockouts                              % This command is only needed if 
\newcommand{\etal}{\textit{et al}.  {\,}}
\newcommand{\ie}{\textit{i}.\textit{e}.}
\newcommand{\PF}{\textit{PointFusion}}
\newcommand{\VF}{\textit{VoxelFusion}}

\title{\LARGE \bf
MVX-Net: Multimodal VoxelNet for 3D Object Detection
}

\author{Vishwanath A. Sindagi$^{1}$, Yin Zhou$^{2}$ and Oncel Tuzel$^{2}$% <-this % stops a space
\thanks{$^{1}$Vishwanath A. Sindagi is with the  Department of Electrical and Computer Engineering, Johns Hopkins University, Baltimore.
        {\tt\small vishwanathsindagi@jhu.edu }}%
\thanks{$^{2}$Yin Zhou and Oncel Tuzel are with AI Research, Apple Inc.
        {\tt\small yzhou3@apple.com, otuzel@apple.com}}%       
}

\begin{document}

\maketitle
\thispagestyle{empty}
\pagestyle{empty}

%%%%%%%%%%%%%%%%%%%%%%%%%%%%%%%%%%%%%%%%%%%%%%%%%%%%%%%%%%%%%%%%%%%%%%%%%%%%%%%%
\begin{abstract}

Many recent works on 3D object detection have focused on designing neural network architectures that can consume point cloud data. While these approaches demonstrate encouraging performance, they are typically based on a single modality and are unable to leverage information from other modalities, such as a camera. Although a few approaches fuse data from different modalities, these methods either use a complicated pipeline to process the modalities sequentially, or perform late-fusion and are unable to learn interaction between different modalities at early stages. In this work, we present \textit{PointFusion} and \textit{VoxelFusion}: two simple yet effective early-fusion approaches to combine the RGB and point cloud modalities, by leveraging the recently introduced VoxelNet architecture. 
Evaluation on the KITTI dataset demonstrates significant improvements in performance over approaches which only use point cloud data. Furthermore, the proposed method provides results competitive with the state-of-the-art multimodal algorithms, achieving top-2 ranking in five of the six bird’s eye view and 3D detection categories on the KITTI benchmark, by using a simple single stage network.

% Furthermore, the proposed method outperforms existing multimodal methods by a large margin.
%In this work, we attempt to fuse information from multiple modalities through two simple techniques: point-fusion and voxel-fusion and are built on the recently proposed Voxel-Net.

\end{abstract}

%%%%%%%%%%%%%%%%%%%%%%%%%%%%%%%%%%%%%%%%%%%%%%%%%%%%%%%%%%%%%%%%%%%%%%%%%%%%%%%%
\section{INTRODUCTION}

With the advent of 3D sensors and diverse applications of 3D understanding, there is an increased research focus on 3D recognition \cite{xiang2016objectnet3d}, object detection \cite{REF:zhou2017voxelnet,REF:qi2017frustum}, and segmentation \cite{REF:qi2017pointnet}. A wide variety of applications such as augmented reality \cite{REF:AR}, robotics \cite{REF:Housekeeping}, and navigation \cite{REF:Geiger2012CVPR,REF:Drone} rely heavily on 3D understanding. Among these tasks, 3D object detection is a fundamental problem and forms a crucial step in many 3D understanding pipelines. In this work, we focus on improving the 3D detection performance by fusing multiple modalities. 

2D object detection is an extensively researched topic in the computer vision community. Convolutional neural network (CNN) based techniques \cite{girshick2015fast,liu2016ssd,lin2017feature,redmon2016you} have shown excellent performance on image-based detection datasets \cite{lin2014microsoft,everingham2010pascal,imagenet_cvpr09}. However, these methods cannot be applied directly to 3D detection since the input modalities are fundamentally different. LiDAR enables accurate localization of objects in the 3D space, and detection techniques based on LiDAR data often outperform the 2D techniques. Some of these methods convert 3D point cloud to hand-crafted feature representations, such as depth or bird's eye view (BEV) maps \cite{REF:cvpr17chen,Yang2018CVPR} and then apply 2D-CNN based methods for vehicle detection and classification. However, these techniques suffer from quantization which leads to reduced performance for objects with fewer points or variable geometries. Another set of techniques represent 3D point cloud data in a voxel grid \cite{REF:maturana_iros_2015,REF:maturana_icra_2014} and employ 3D CNNs to generate detection results. These methods are often limited by the memory requirements, especially when processing full scenes.

\begin{figure}[!t]
	\centering
	\includegraphics[width=1\linewidth]{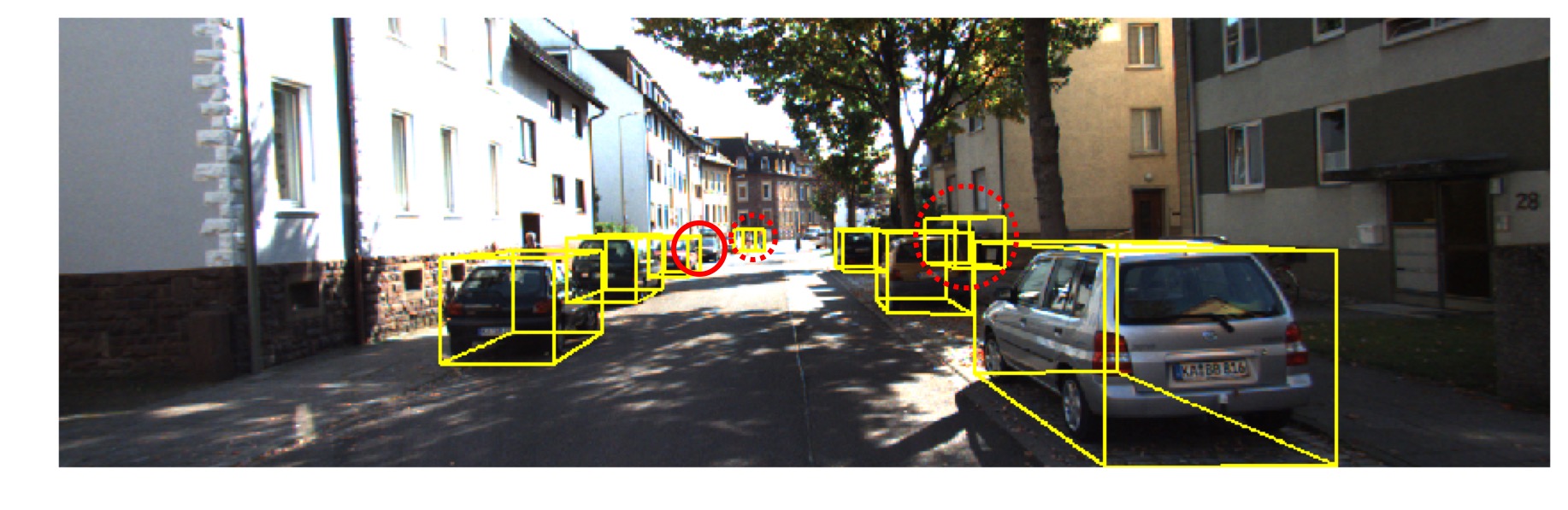}
	\includegraphics[width=1\linewidth]{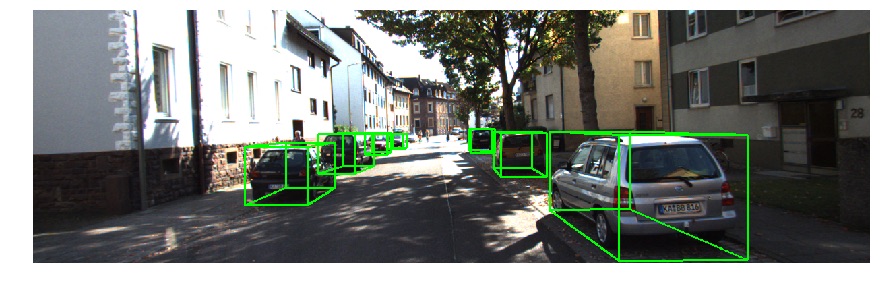}	
	\vskip-15pt
	\caption{Example 3D detection result from the KITTI validation set projected onto an image. \textit{Top row:} VoxelNet \cite{REF:zhou2017voxelnet}, where yellow boxes represent detections. The solid red circle and dashed red circles highlight a false negative and two false positives by VoxelNet, respectively. \textit{Bottom row:} Proposed method, where green rectangles indicate detections.}
	\label{fig:sampleresult}
	\vspace{-0.5cm}
\end{figure}

\begin{figure*}[ht!]
	\centering
	\includegraphics[width=.75\linewidth]{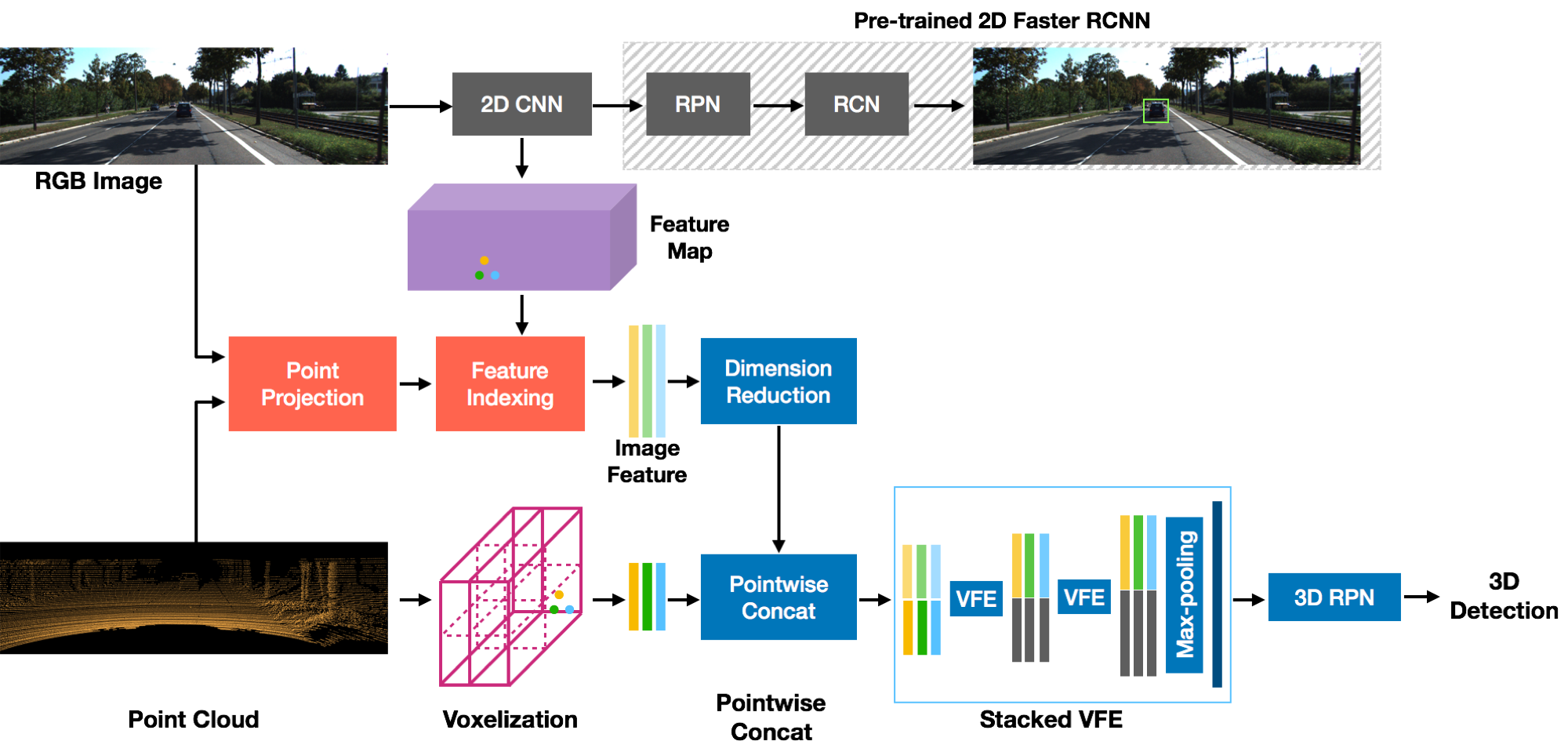}
	\vskip-10pt	
	\caption{Overview of the proposed MVX-Net \PF{\,}method. The method uses convolutional filters of pre-trained 2D faster RCNN to compute the image feature map. Note that the RPN and RCN (shown in shaded rectangle) are not part of the 3D inference pipeline. The 3D points are projected to the image using the calibration information and the corresponding image features are appended to the 3D points. The VFE layers and the 3D RPN process the aggregated data and produce the 3D detections. }
	\label{fig:pfarch}
	\vspace{-0.5cm}
\end{figure*}

Recent research on 3D classification has focused on enabling the use of end-to-end trainable neural networks that can consume point cloud data without transforming them to intermediate representations, such as depth or BEV formats. Qi \etal \cite{REF:qi2017pointnet} designed a neural network architecture that directly takes point clouds as input and outputs class labels. With this design, one can learn the representations from the raw data. However, this work could not be applied to the problem of detection and localization due to the limitations in architecture design along with high computational and memory cost. Recently, Zhou and Tuzel \cite{REF:zhou2017voxelnet} overcame this issue by proposing VoxelNet, which involves voxelization of a point cloud and encoding the voxels using stacks of Voxel Feature Encoding (VFE) layers. With these steps, VoxelNet enabled the use of a 3D region proposal network for detection. Although this method demonstrates encouraging performance, it relies on a single modality, \ie, point cloud data. In contrast to point clouds, RGB images provide much denser texture information and it is desirable to leverage both modalities to improve the detection performance. 

In this paper, we propose Multimodal VoxelNet (MVX-Net), to augment LiDAR points with semantic image features and learn to fuse image and LiDAR features at early stages for accurate 3D object detection. The proposed approach extends the recently proposed VoxelNet algorithm \cite{REF:zhou2017voxelnet}. Specifically, we develop two fusion techniques: (i) \PF: This is an early-fusion method where points from the LiDAR sensor are projected onto the image plane, followed by image feature extraction from a pre-trained 2D detector. The concatenation of image features and the corresponding points are then jointly processed by the VoxelNet architecture. (ii) \VF: In this technique, non-empty 3D voxels created by VoxelNet are projected to the image, followed by extracting image features for every projected voxel using a pre-trained CNN. These features are then pooled and appended to the VFE feature encoding for every voxel and further used by the 3D region proposal network (RPN) to produce 3D bounding boxes. Compared to \PF, \VF{\,}is a relatively later fusion technique, which however, can be extended to handle empty voxels as well, thereby reducing the dependency on the availability of high-resolution LiDAR point cloud data. As illustrated in Fig. \ref{fig:sampleresult}, the proposed MVX-Net effectively fuses multimodal information leading to reduced false positives and negatives compared to the LiDAR-only VoxelNet.

This paper is organized as follows. Section~\ref{sec:relatedwork} describes related work on 3D object detection. Section~\ref{sec:proposedmethod} introduces the proposed Multimodal VoxelNet algorithm and two fusion techniques for effectively combining multimodal information. Section~\ref{sec:experiments} presents experimental results. Finally, Section~\ref{sec:conclusion} concludes the paper and points out future directions for improvement.

%Next, we briefly review related work (Section \ref{sec:relatedwork}) and provide details of the proposed method in Section \ref{sec:proposedmethod} with experiments and results in Section \ref{sec:experiments}, followed by conclusions in Section \ref{sec:conclusion}.

\section{RELATED WORK}
\label{sec:relatedwork}
As discussed earlier, 3D understanding is an extensively researched topic. Earlier approaches (\cite{REF:PointSignatures_Chua1997,REF:StructuralIndexing_Medioni92,REF:Nishino2010,REF:COSMOS_Jain97,REF:Tuzel2014}) employ hand-crafted representations and achieve satisfactory results in the presence of rich and detailed 3D information. Some of the 	techniques (\cite{REF:Wang-RSS-15,REF:Vote3Deep,REF:SlidingShapes_Song2014,REF:SlidingShapes_Song2014,REF:3DFCN}) represent 3D point cloud data using voxel occupancy grid representation, followed by the use of 3D convolutions to compute the 3D bounding boxes. Due to the high computational and memory cost, several approaches based on BEV representation were developed (\cite{REF:FusionDPM-IROS14,REF:MV-RGBD-RF2015,Yang2018CVPR}). The BEV-based methods assume that point cloud data is sparse in one dimension, which is usually not the case in many scenarios. Different from these approaches, image-based methods (\cite{REF:cvpr16chen,REF:nips15chen,REF:xiang_cvpr15,REF:Zia2013,REF:Zia2014,REF:SFM2015,REF:chabot2017deep}) were developed to infer 3D bounding boxes from 2D images. However, they usually suffer from low accuracy in terms of depth localization. Recently, VoxelNet \cite{REF:zhou2017voxelnet} proposed an end-to-end learning architecture that consumes point cloud data in its raw format. 

Multimodal fusion by combining LiDAR and RGB data has been less explored as compared to single modality-based approaches. Recently, Chen \etal \cite{REF:cvpr17chen} proposed a multi-view 3D object detection network (MV3D), which takes multimodal data as input and produces 3D bounding boxes by incorporating region-based feature fusion. Although this method demonstrated encouraging results by using multimodal data, it has the following disadvantages: (i) the method converts point clouds into BEV representation, which loses detailed 3D shape information, and (ii) the fusion is performed at a much later stage as compared to the proposed fusion techniques (\ie, after the 3D proposal generation stage), which limits the ability of the neural network to capture the interaction between the two modalities at earlier stages and hence, the integration is not necessarily seamless. Similar to \cite{REF:cvpr17chen}, Ku \etal \cite{ku2017joint} proposed multimodal fusion by incorporating region-based features. They achieved better performance than \cite{REF:cvpr17chen} especially in the small object category by designing a more advanced RPN that employs high-resolution feature maps. This method also uses hand-crafted BEV representation and performs late fusion. 

In a different approach, Qi \etal proposed Frustum PointNets \cite{REF:qi2017frustum} for 3D detection using LiDAR and RGB data. First, they use a 2D object detector on RGB data to generate 2D proposals, which are then converted to frustum proposals in the 3D space, followed by a point-wise instance segmentation using the PointNet architecture \cite{REF:qi2017pointnet}. This method is an image-first  approach and hence lacks the capability of utilizing both modalities simultaneously. Most recently, Liang \etal \cite{Liang_2018_ECCV} proposed to aggregate the discrete BEV space with image features by projecting the LiDAR points to image space. This approach interpolates each BEV pixel location with RGB features based on K nearest neighbor search, which may not satisfy real-time requirements as the density and coverage of LiDAR point clouds increase. 
In contrast to existing approaches that either use a complicated pipeline to process different modalities or perform late-fusion, our simple yet effective fusion strategies can learn interaction between modalities at early stages.

% TODO TODO TODO TODO TODO TODO TODO TODO TODO TODO TODO TODO TODO TODO TODO
% Add avod and continuous fusion

% TODO TODO TODO TODO TODO TODO TODO TODO TODO TODO TODO TODO TODO TODO TODO

%\section{RELATED WORK}

%\subsection{3D Object Detection}
%\subsection{Multimodal Fusion}
%\clearpage

\begin{figure*}[ht!]
	\centering
	\includegraphics[width=.75\linewidth]{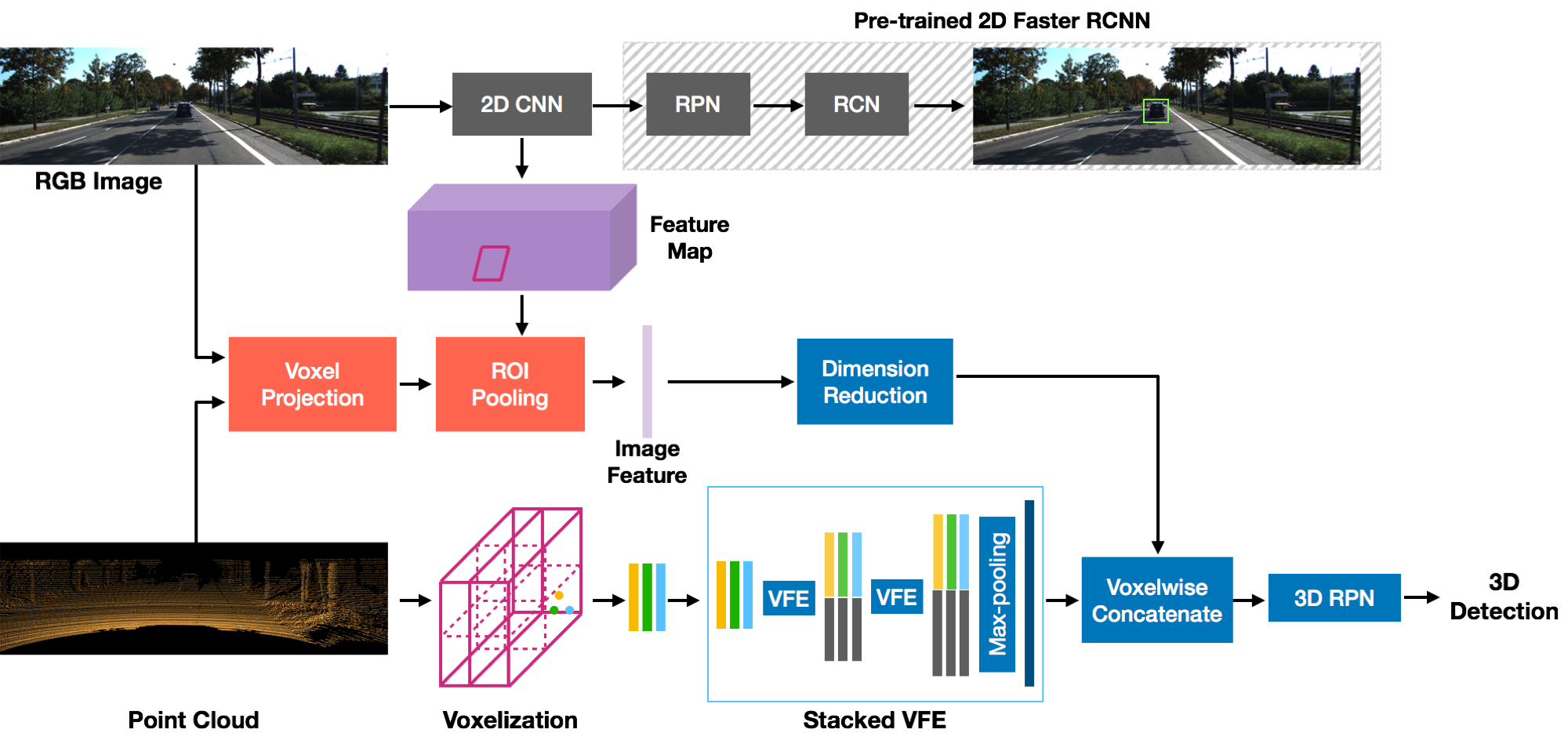}
	\vskip-10pt	
	\caption{Overview of the proposed MVX-Net \VF{\,}method. The method uses convolutional filters of pre-trained 2D faster RCNN to compute the image feature map. Note that the RPN and RCN (shown in shaded rectangle) are not part of the 3D inference pipeline. The non-empty voxels are projected to the image using the calibration information to obtain the ROIs. The features within each ROI are pooled and appended to the voxel features computed by VFE layers. The 3D RPN processes the aggregated data and produces the 3D detections.}
	\label{fig:vfarch}
	\vspace{-0.5cm}
\end{figure*}

\section{PROPOSED METHOD}
\label{sec:proposedmethod}
The proposed fusion techniques, illustrated in Fig.~\ref{fig:pfarch} and Fig.~\ref{fig:vfarch}, are based on the VoxelNet~\cite{REF:zhou2017voxelnet} architecture. In order to fuse information from RGB and point cloud data, we first extract features from the last convolutional layer of a 2D detection network. This network is first pre-trained on ImageNet~\cite{REF:ILSVRC15,simonyan2014very} and then fine-tuned for the 2D object detection~\cite{ren2015faster} task. These high-level image features encode semantic information that can be used as prior knowledge to help infer the presence of an object. Based on the type of fusion described earlier (\PF{\,}or \VF), either points or voxels are projected onto the image and the corresponding features are concatenated with point features or voxel features respectively. Details of the 2D detection network, VoxelNet, and the proposed fusion techniques are described in the following subsections.

\subsection{2D Detection Network}
\label{ssec:2ddetector}

Compared to LiDAR point clouds, RGB images capture richer color and texture information. In this work, to improve 3D detection accuracy, we extract high-level semantic features from RGB images and incorporate them into the VoxelNet algorithm.

Convolutional neural networks are highly effective at learning semantic information present in the images. Here, we propose to use an existing 2D detection framework which has shown excellent performance on various tasks \cite{lin2014microsoft,everingham2010pascal,imagenet_cvpr09}. Specifically, we employ the Faster-RCNN framework \cite{ren2015faster} which consists of a region proposal network (RPN) and a region classification network (RCN). We use VGG16 \cite{simonyan2014very} pre-trained on ImageNet~\cite{REF:ILSVRC15} as the base network and finetune the Faster-RCNN network using images from a 2D detection dataset and the corresponding bounding box annotations. More training details are described in Section \ref{ssec:training}.

Once the detection network is trained, high-level features (from the conv5 layer of the VGG16 network) are extracted and fused either at the point or voxel level.

\subsection{VoxelNet}

We choose the VoxelNet architecture as the base 3D detection network for two main reasons: (i) it consumes raw point clouds and removes the need for hand-crafted features and (ii) it provides a natural and effective interface for combining image features at different granualities in 3D space, \textit{i.e.,} points and voxels. 
We use the network as described in \cite{REF:zhou2017voxelnet}. For completeness, we briefly revisit VoxelNet in this section. This algorithm consists of three building blocks: (i) a Voxel Feature Encoding (VFE) layer (ii) Convolutional Middle Layers, and (iii) a 3D Region Proposal Network. 

VFE is a feature learning network that aims to encode raw point clouds at the individual voxel level. Given a point cloud, the 3D space is divided into equally spaced voxels, followed by grouping the points to voxels. Then each voxel is encoded using a hierarchy of voxel feature encoding layers. First, every point $\mathbf{p}_i=[x_i, y_i, z_i, r_i]^{T}$ (containing the XYZ coordinates and the reflectance value) in a voxel is represented by its co-ordinates and its relative offset with respect to the centroid of the points in the voxel. That is each point is now represented as: $ \hat{\mathbf{p}}_i=[x_i, y_i, z_i, r_i, x_i-v_x, y_i-v_y, z_i-v_z]^{T} \in \mathbb{R}^7$, where $x_i, y_i, z_i, r_i$ are the XYZ coordinates and the reflectance value of the point $p_i$, and $v_x, v_y, v_z$ are the XYZ coordinates of the centroid of the points in the voxel which $p_i$ belongs to. Next, each $\hat{\mathbf{p}}_i$ is transformed through the VFE layer which consists of a fully connected network (FCN) into a feature space, where information from the point features can be aggregated to encode the shape of the surface contained within the voxel. The FCN is composed of a linear layer, a batch normalization (BN) layer, and a rectified linear unit (ReLU) layer. The transformed features belonging to a particular voxel are aggregated using element-wise max-pooling. The max-pooled feature vector is then concatenated with point features to form the final feature embedding. All non-empty voxels are encoded in the same way and they share the same set of parameters in FCN. Stacks of such VFE layers are used to transform the input point cloud data into high-dimensional features. 

The output of the stacked VFE layers are forwarded through a set of convolutional middle layers that apply 3D convolution to aggregate voxel-wise features within a progressively expanding receptive field. These layers incorporate additional context, thus enabling the use of context information to improve the detection performance.

Following the convolutional middle layers, a region proposal network \cite{girshick2015fast} performs the 3D object detection. This network consists of three blocks of fully convolutional layers. The first layer of each block downsamples the feature map by half via a convolution with a stride size of 2, followed by a sequence of convolutions of stride 1. After each convolution layer, BN and ReLU operations are applied. The output of every block is then upsampled to a fixed size and concatenated to construct a high resolution feature map. Finally, this feature map is mapped to the targets: (1) a probability score map and (2) a regression map.

\subsection{Multimodal Fusion}
\label{ssec:mfusion}

As discussed earlier, VoxelNet \cite{REF:zhou2017voxelnet} is based on a single modality and adapting it to multimodal input enables further performance improvements. In this paper, we propose two simple techniques to fuse RGB data with the point cloud data by extending the VoxelNet framework. 

\noindent\textbf{\PF}: This is an early fusion technique where every 3D point is aggregated by an image feature to capture a dense context.

The method first uses a pre-trained 2D detection network (described in Section \ref{ssec:2ddetector}) to extract a high level feature map from the image which encodes image-based semantics. Then using the calibration matrix, it projects each 3D point onto the image and appends the point with the feature corresponding to the projected location index. This process associates information about the presence and, if it exists, the pose of the object from 2D images to every 3D point. Note that the features are extracted from the conv5 layer of the VGG16 network and are 512 dimensional. We first reduce the dimensionality to 16 through a set of fully connected layers and then concatenate them to the point features. The concatenated features are processed by a set of VFE layers in VoxelNet and then used in the detection stage. Fig.~\ref{fig:pfarch} provides an overview of this approach. 

The advantage of this approach is that since the image features are concatenated at a very early stage, the network can learn to summarize useful information from both modalities through the VFE layer. Moreover, the approach exploits the LiDAR point cloud and lifts the corresponding image features to the coordinates of the 3D points.
 % is simple to implement and use as it consists of only two extra steps \ie projection of points to image and extraction of high-level features from a pre-trained CNN. However, this approach is still LiDAR-centric, in the sense that availability high-quality LiDAR data is still crucial to obtain high quality detections. \\

\noindent\textbf{\VF}: In contrast to \PF{\,}that combines features at an earlier stage, \VF{\,}employs a relatively later fusion strategy where the features from the RGB image are appended at the voxel level. As described in \cite{REF:zhou2017voxelnet}, the first stage in VoxelNet involves dividing the 3D space into a set of equally spaced voxels. Points are grouped into these voxels based on where they reside, after which each voxel is encoded using a VFE layer. In the proposed \VF{\,}method, every non-empty voxel is projected onto the image plane to produce a 2D region of interest (ROI). Using the feature map from the pre-trained detector network (conv5 layer of VGG16), the features within the ROI are pooled to produce a 512-dimensional feature vector, whose dimensionality is first reduced to 64 and then appended to the feature vector produced by the stacked VFE layers at every voxel. This process encodes prior information from the 2D image at every voxel. Fig.~\ref{fig:vfarch} provides an overview of this approach.

Although \VF{\,}is a relatively later fusion strategy and produces slightly inferior performance as compared to \PF, it has the following advantages. First, it can be easily extended to aggregate image information to empty voxels where LiDAR points are not sampled due to reasons such as low LiDAR resolution or far objects, thereby reducing dependency on the availability of high-resolution LiDAR points. Second, \VF{\,}is more efficient in terms of memory consumption as compared to \PF.

%Compared to the \PF, this technique fuses information from the two modalities at a later stage. %Due to this reason, the network is unable to capture.. (TODO). 
%It can be observed from Section \ref{sec:experiments}, that although this method improves the performance of VoxelNet, it is slightly inferior to the \PF technique. In contrast to \PF which is LiDAR-centric, this technique can be easily extended to use both modalities equally. In our current implementation, we project only non-empty voxels to the image before the feature extraction. The method can be easily extended  by projecting all voxels onto the image and considering only those projected voxels that have sufficient overlap with the 2D detections. By doing this, we will be encoding information into those voxels where point cloud data is missing. This will be especially helpful in improving the detection performance of far-range objects where point cloud data has relatively lower resolution. With image-based detectors being more evolved \cite{girshick2015fast,liu2016ssd,lin2017feature,redmon2016you}, they perform exceptionally well at detecting small objects and can be easily leveraged in the \VF framework to build a 3D detection network that relies on both modalities equally. 

\subsection{Training Details}
\label{ssec:training}
% In this section, we describe details of the training procedure for the 2D detector and multimodal VoxelNet. 

\noindent\textbf{2D Detector}: We use the standard Faster-RCNN detection framework \cite{ren2015faster}, which is a two stage detection pipeline consisting of a region proposal network and a region classification network. The base network is VGG16 architecture and we use ROIAlign \cite{he2017mask} operation to pool the features from the last convolutional layer before forwarding them to the second stage (RCNN). We use four sets of anchors with sizes \{4,8,16,32\} and three aspect ratios \{0.5,1,2\} on the conv5 layer. Anchors are labeled as positive if the intersection-over-union (IoU) with the ground truth boxes is greater than 0.7, and the anchors are labeled as negative if the IoU is less than 0.3. During training, the shortest side of the image is rescaled to 600 pixels. The training dataset is augmented with standard techniques such as flipping and adding random noise. For the RCNN stage, we use a batch size of 128 with 25\% of the samples reserved for foreground ROIs. The network is trained using stochastic gradient descent with a learning rate of 0.0005 and momentum of 0.9.

\noindent\textbf{Multimodal VoxelNet}: We retain most of the settings of VoxelNet as described in \cite{REF:zhou2017voxelnet} apart from a few simplifications to improve the efficiency. 
The 3D space is divided into voxels of sizes $v_D=0.4$, $v_H=0.2$, $v_W=0.2$. Two sets of VFE layers and three convolutional middle layers are used. The input and output dimensionalities of these layers are different based on the type of fusion. 

For \PF, the VFE stack has a configuration of VFE-1(7+16,32) and VFE-2(32,128). The input to the first VFE layer is a concatenation of point features which have 7 dimensions and CNN features which have 16 dimensions. Note that the features extracted from conv5 layer of the pre-trained 2D detection network have a dimensionality of 512. Their dimensions are first reduced to 96 and finally to 16 using two fully-connected (FC) layers with BN and ReLU. 

For \VF, the VFE stack has a configuration of VFE-1(7,32) and VFE-2(32,64). Features extracted from conv5 layer of the pre-trained 2D detection network have a dimensionality of 512, and they are reduced to 128D and 64D using two FC layers, each followed by a BN and a ReLU non-linearity. These dimension reduced features are then concatenated to the output of VFE-2 to form a 128 dimensional vector for every voxel. By reducing the output dimensionality of VFE-2 to 64 (as compared to 128 in the original work), we ensure that the architecture of the convolutional middle layers remain unchanged. 

To reduce the memory footprint, we trim the RPN by using only half of the number of ResNet blocks as in the original work. We employ the same anchor matching strategies as those in the original work. For both fusion techniques, the network is trained using stochastic gradient descent with a learning rate of 0.01 for the first 150 epochs, after which the learning rate is decayed by a factor of 10. Furthermore, since we use both images and point clouds, some of the augmentation strategies used in the original work are not applicable to the proposed multimodal framework, e.g., global point cloud rotation. Despite training with a trimmed RPN and using less data augmentations, the proposed multimodal framework is still able to achieve significantly higher detection accuracy compared to the original LiDAR-only VoxelNet~\cite{REF:zhou2017voxelnet}.

\begin{table}[t!]
	\caption{\scriptsize COMPARISON OF RESULTS ON KITTI VALIDATION SET USING MEAN AVERAGE PRECISION (in \%) WITH IOU=0.7. Top-2 methods are highlighted in bold. (S: SINGLE MODALITY, M:MULTIMODAL)}
	\label{tab:results70}
	\centering
	\resizebox{0.9\columnwidth}{!}{%
	\begin{tabular}{|l|c|c|c|c|c|c|}
		\hline
		\multirow{2}{*}{Method}                  & \multicolumn{3}{c|}{AP$_{BEV}$ (IoU=0.7)}           & \multicolumn{3}{c|}{AP$_{3D}$ (IoU=0.7)}            \\ \cline{2-7} 
		& Easy           & Med       & Hard           & Easy           & Med       & Hard  \\ \hline
		Mono3D \cite{REF:cvpr16chen} (S)         & 5.22           & 5.19           & 4.13           & 2.53           & 2.31           & 2.31  \\ \hline
		3DOP \cite{REF:nips15chen} (S)           & 12.6          & 9.49           & 7.5           & 6.55           & 5.07           & 4.10  \\ \hline
		VeloFCN \cite{REF:VeloFCN} (S)           & 40.1          & 32.0          & 30.4          & 15.2           & 13.6          & 15.9 \\ \hline
		MV3D \cite{REF:cvpr17chen} (S)           & 86.2          & 77.3          & 76.3          & 71.2          & 56.6          & 55.3 \\ \hline
		MV3D \cite{REF:cvpr17chen} (M)           & 86.6          & 78.1          & 76.7          & 71.3          & 62.7          & 56.6 \\ \hline
		PIXOR \cite{Yang2018CVPR} (S)            & 86.8          & 80.8          & 76.6          & N/A           & N/A           & N/A \\ \hline
		F-PointNet \cite{REF:qi2017frustum} (M)       & 88.2          & 84.0          & 76.4          & \textbf{83.8}          & 70.9          & 63.7 \\ \hline
		VoxelNet \cite{REF:zhou2017voxelnet} (S) & \textbf{89.6} & \textbf{84.8}          & \textbf{78.6}          & 82.0          & 65.5          & 62.9 \\ \hline 
		Baseline VoxelNet  (S) & 87.6 & 83.7          & 78.4          & 79.5          & 65.7          & 64.6 \\
		MVX-Net (VF) (M)                         & 88.6          & 84.6          & \textbf{78.6}          & 82.3          & \textbf{72.2}          & \textbf{66.8} \\ %\hline
		%MVX-Net (PF2) (M)                     & 89.0          & 84.7          & 78.8          & 85.0          & \textbf{73.3}          & 67.2 \\ %\hline
		MVX-Net (PF) (M)                     & \textbf{89.5}          & \textbf{84.9} & \textbf{79.0} & \textbf{85.5} & \textbf{73.3} & \textbf{67.4} \\ \hline
	\end{tabular}
}
% \vspace{-0.5cm}
\end{table}

\begin{table}[t!]
	\caption{\scriptsize COMPARISON OF RESULTS ON KITTI VALIDATION SET USING MEAN AVERAGE PRECISION (MAP) WITH IOU=0.8.}
	\label{tab:results80}
	\centering
	\resizebox{0.9\columnwidth}{!}{%
	\begin{tabular}{|l|c|c|c|c|c|c|}
		\hline
		\multirow{2}{*}{Method}                  & \multicolumn{3}{c|}{AP$_{BEV}$ (IoU=0.8)} & \multicolumn{3}{c|}{AP$_{3D}$ (IoU=0.8)} \\ \cline{2-7} 
		& Easy            & Med            & Hard           & Easy            & Med           & Hard           \\ \hline
		Baseline VoxelNet  (S) & 72.4         & 62.2         & 56.5        & 32.8         & 28.1        & 24.6        \\ \hline
		MVX-Net (VF) (M)                         & 72.2         & 62.3         & 61.0        & 39.5         & 30.8        & 29.8        \\ \hline
		MVX-Net (PF) (M)                     & 74.2       & 64.5         & 61.6        & 43.6         & 33.2       & 31.3        \\ \hline
	\end{tabular}
	}
\end{table}

\begin{figure*}[t!]
	\centering
	\includegraphics[width=0.31\linewidth]{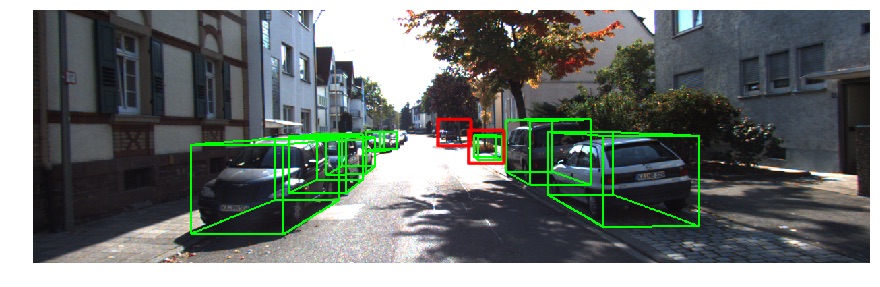}
	\includegraphics[width=0.31\linewidth]{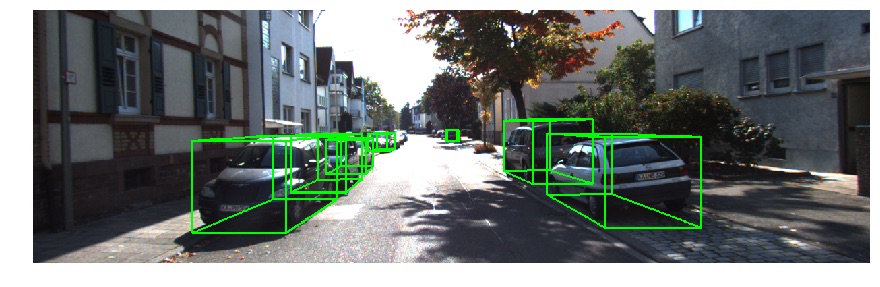}	
	\includegraphics[width=0.31\linewidth]{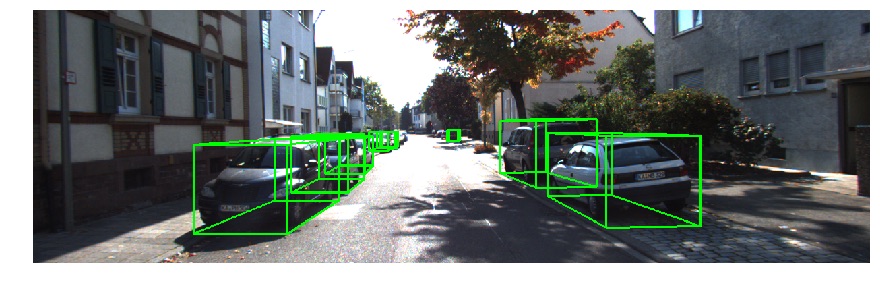}	
	
	\includegraphics[width=0.31\linewidth]{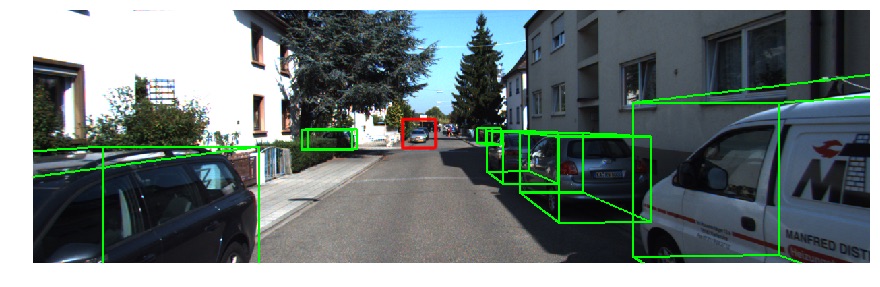}
	\includegraphics[width=0.31\linewidth]{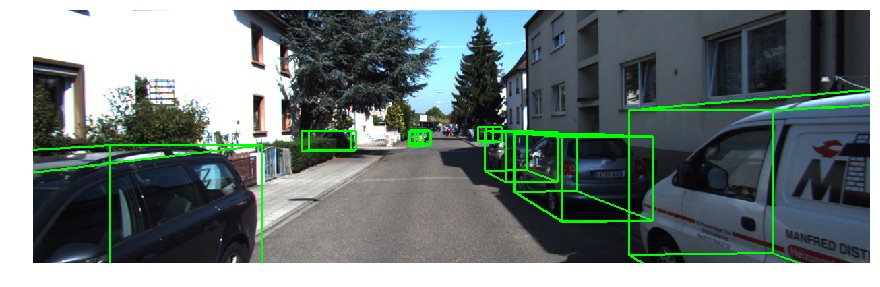}	
	\includegraphics[width=0.31\linewidth]{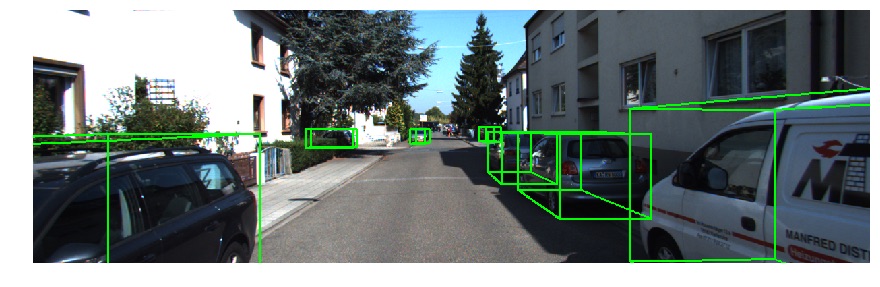}	
	
	\includegraphics[width=0.31\linewidth]{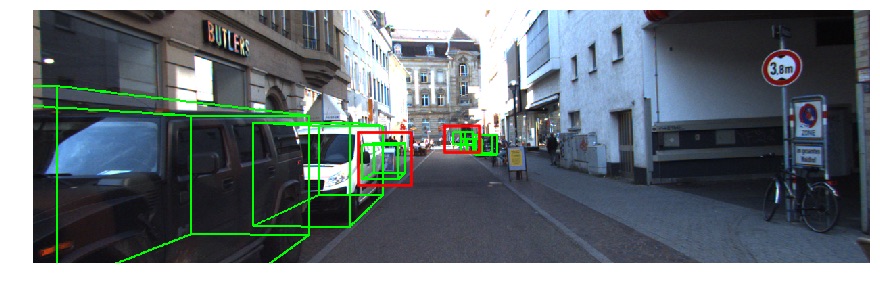}
	\includegraphics[width=0.31\linewidth]{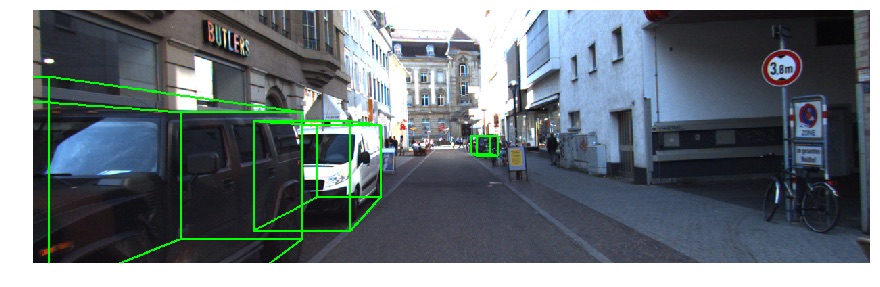}	
	\includegraphics[width=0.31\linewidth]{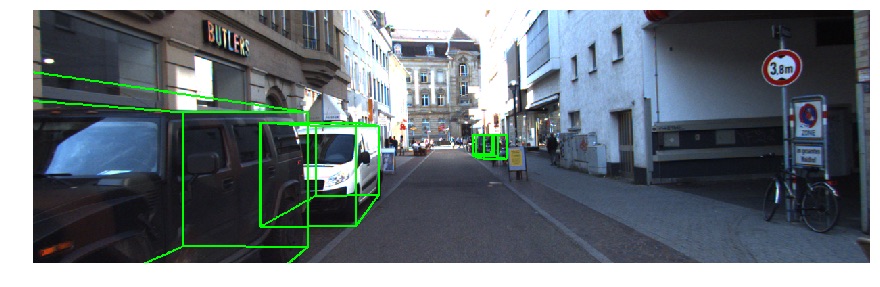}	
	
	\includegraphics[width=0.31\linewidth]{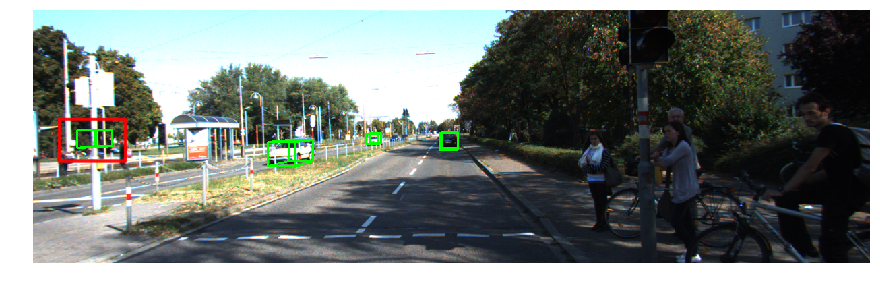}
	\includegraphics[width=0.31\linewidth]{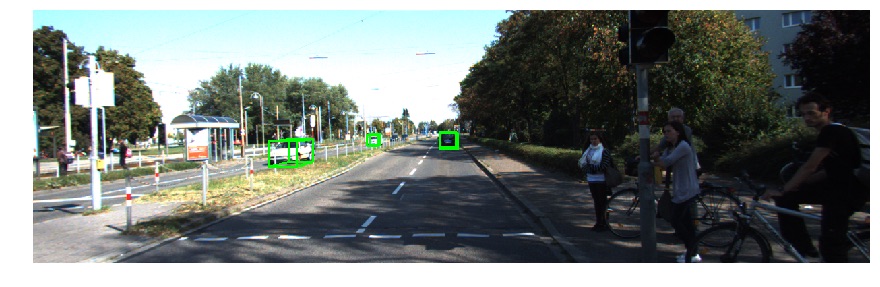}	
	\includegraphics[width=0.31\linewidth]{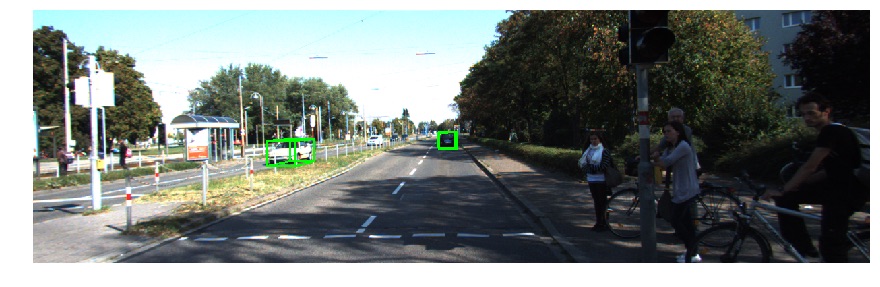}	
	
	\includegraphics[width=0.31\linewidth]{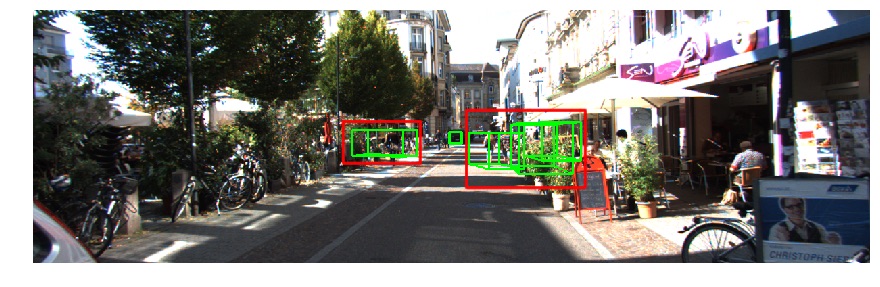}
	\includegraphics[width=0.31\linewidth]{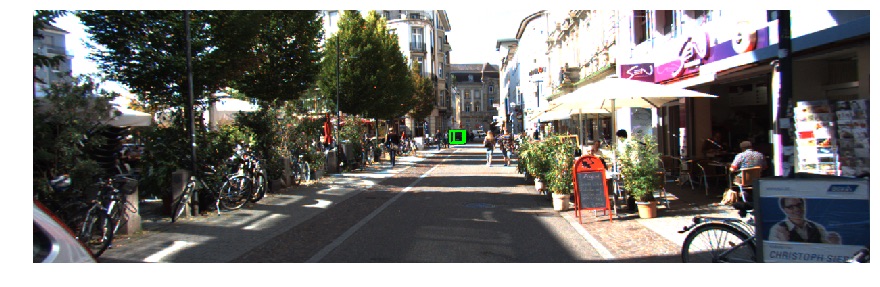}	
	\includegraphics[width=0.31\linewidth]{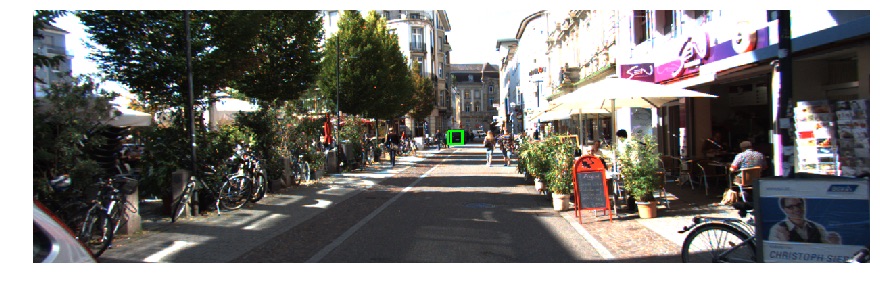}	

	(a)\hskip150pt (b)\hskip150pt (c)
	\vskip-0pt
	\caption{Sample 3D detection results from KITTI validation dataset projected onto image for visualization. \textit{(a)} VoxelNet \cite{REF:zhou2017voxelnet}, \textit{(b)} MVX-Net with \textit{VoxelFusion}, \textit{(c)} MVX-Net with \textit{PointFusion}. Green rectangles indicate detection results. Red rectangles highlight missed detections and false positives.}
	\label{fig:results_image}
\end{figure*}

\section{EXPERIMENTS AND RESULTS}
\label{sec:experiments}
\subsection{Dataset}
\label{ssec:Dataset}
The proposed fusion techniques are evaluated on the KITTI 3D object detection dataset \cite{REF:Geiger2012CVPR} that contains 7,481 training samples and 7,518 test samples. 
%The dataset consists of three categories of objects: Car, Pedestrian and Cyclist and these categories 
There are three difficulty levels: easy, moderate and hard which are determined based on the object size, visibility (occlusion) and truncation. We further split the training set into train/validation sets by avoiding samples from the same sequence being included in both sets \cite{REF:cvpr17chen}. After the split, the training set consists of 3712 samples and the validation set consists of 3769 samples. 

We compare the proposed MVX-Net with previously published approaches on the car detection tasks. To analyze effectiveness of the proposed multimodal approaches we also trained a baseline VoxelNet model. Similar to the multimodal approaches, this model used the trimmed architecture and did not use the global rotation augmentation. By comparing the results to this baseline, we can directly attribute the gains to the proposed multimodal fusion techniques. \\

\subsection{Evaluation on KITTI Validation Set}
\label{ssec:validationset}
We follow the standard KITTI evaluation protocol (IoU=0.7) for measuring the detection performance. Table~\ref{tab:results70} shows the mean average precision (mAP) scores for \VF{\,}and \PF{\,}compared to the state-of-the-art methods on the KITTI validation set using 3D and bird's eye view (BEV) evaluation. In all fusion experiments, the detection performance improved significantly after fusion as compared to the baseline VoxelNet. The effectiveness of fusion is more pronounced in terms of 3D mAP scores than BEV mAP scores. It is also important to note that the proposed fusion techniques are able to obtain improved performance as compared to the original VoxelNet which has a more powerful RPN and uses more data augmentations. Moreover, our approach consistently outperforms other recent top-performing approaches~\cite{REF:cvpr17chen,Yang2018CVPR,REF:qi2017frustum}. Fig.~\ref{fig:results_image} compares example detection results from the proposed approaches and the LiDAR-only VoxelNet~\cite{REF:zhou2017voxelnet}. 

It can be observed that \VF{\,}yields slightly lower performance as compared to \PF{\,}because \PF{\,}combines features at an earlier stage. It is worth pointing out that in contrast to \PF{\,}which is LiDAR-centric, \VF{\,}can exploit both modalities independently. For efficient training and inference, our current implementation only projects non-empty voxels onto the image. However, the \VF{\,}method can be extended by projecting all voxels onto the image. This strategy utilizes image-based information regardless of existence of the points within a voxel which could be helpful in far-range detection, where LiDAR has a very low resolution.

We conduct an ablation study by replacing the 2D CNN features by the cropped raw image patches and append them to the 3D points. We test image patch sizes 3x3 and 5x5 and find that even this simple strategy yields 0.5\% to 1.0\% higher AP as compared to the baseline VoxelNet model. However, this result is not as good as using a higher level feature computed through an image CNN.

% Fig. \ref{fig:pr70} shows the precision-recall (PR) curves for different fusion techniques along with the original VoxelNet. It can be observed from the top row of this figure that the fusion techniques improve both precision and recall in all the categories. While the mAP scores in the BEV evaluation do not reveal much difference in the performance after fusion, the precision-recall curves in the bottom row of Fig. \ref{fig:pr70} provide further insight into the effect of fusion. 

% It can also be observed that fusion improves the precision/recall especially in the moderate and hard categories, with \PF performing better than other techniques in all categories. 

\begin{table}[t!]
	\caption{\scriptsize COMPARISON OF RESULTS ON KITTI TEST SET USING MEAN AVERAGE PRECISION (in \%) WITH IOU=0.7. Top-2 methods are highlighted in bold. (S: SINGLE MODALITY, M:MULTIMODAL)}
	\label{tab:test_set_result}
	\centering
	\resizebox{\columnwidth}{!}{%
		\begin{tabular}{|l|c|c|c|c|c|c|}
			\hline
			\multirow{2}{*}{Method}                  & \multicolumn{3}{c|}{AP$_{BEV}$ (IoU=0.7)}           & \multicolumn{3}{c|}{AP$_{3D}$ (IoU=0.7)}            \\ \cline{2-7} 
			& Easy           & Med       & Hard           & Easy           & Med       & Hard  \\ \hline
			MV3D \cite{REF:cvpr17chen}    (S)       & 85.8          & 77.0          & 68.9          & 66.8          & 52.7          & 51.3 \\ \hline
			PIXOR \cite{Yang2018CVPR}  (S)&  81.7  &  77.1  &  73.0  &  N/A  &  N/A  &  N/A   \\ \hline
			PIXOR++ \cite{Yang2018CoRL}  (M) &  \textbf{89.4}  &  83.7  &  78.0  &  N/A  &  N/A  &  N/A   \\ \hline
			VoxelNet \cite{REF:zhou2017voxelnet} (S)  & \textbf{89.4} &  79.3  &  77.4        & 77.5          & 65.1           & 57.7 \\ \hline
			MV3D \cite{REF:cvpr17chen}  (M)           & 86.0          & 76.9          & 68.5          & 71.1          & 62.4          & 55.1 \\ \hline
			F-PointNet \cite{REF:qi2017frustum}  (M)     & 88.7          & 84.0          & 75.3          & 81.2          & 70.4          & 62.2  \\ \hline
			AVOD \cite{ku2017joint} (M)  &  86.8   &   85.4  &   77.7  &  73.6  &  65.8  &  58.4 \\ \hline
			AVOD-FPN \cite{ku2017joint} (M) &  88.5   &   83.8  &   77.9 &  81.9  &  \textbf{71.9}  &  \textbf{66.4} \\ \hline
			HDNET \cite{Yang2018CoRL} (M)  &  89.1  &  \textbf{86.6}  &  \textbf{78.3}  &  N/A  &  N/A  &  N/A   \\ \hline
			Cont-Fuse \cite{Liang_2018_ECCV}  (M)  &  88.8  &  85.8  &  77.3  &  \textbf{82.5}  &  66.2  &  64.0  \\ \hline
			
			MVX-Net (PF) (M)                     & 89.2          & \textbf{85.9} & \textbf{78.1} & \textbf{83.2} & \textbf{72.7}  & \textbf{65.2} \\ \hline
		\end{tabular}
	}
\end{table}

We investigate the performance of different methods with a more rigorous evaluation criterion by increasing the IoU threshold to 0.8. The mAP scores for this configuration are summarized in Table \ref{tab:results80}. 
As the IoU criterion increases, the performance improvement by multimodal fusion is more pronounced, which indicates that multimodal fusion helps to improve not only the detection but also the localization accuracy over approaches using
a single modality. \\

\subsection{Evaluation on KITTI Test Set}
\label{ssec:testset}
We evaluate the proposed MVX-Net with \PF{\,}on the KITTI test set by submitting detection results to the official server. The results are summarized in Table~\ref{tab:test_set_result}. We observe that the MVX-Net with \PF{\,}achieves competitive results with the state-of-the-art 3D detection algorithms. Out of six bird’s eye view and 3D detection categories, the proposed approach achieves top rank in two categories,  2nd rank in three categories, and 3rd in one other category.

% Note that the improvements as compared to the original VoxelNet \cite{REF:zhou2017voxelnet} are significant, especially considering the fact that our approach uses a trimmed RPN and employs fewer augmentations.
%We can see that the proposed approach yields very competitive performance compared with other top-performing algorithms in both the BEV and the 3D car detection tasks. 


\section{CONCLUSION}
\label{sec:conclusion}

In this work, we present two feature fusion techniques, \PF{\,}and \VF, to combine RGB with LiDAR, by extending the recently proposed VoxelNet \cite{REF:zhou2017voxelnet}. \PF{\,}involves projection of 3D points onto the image using a known calibration matrix, followed by feature extraction from a pre-trained 2D CNN and concatenation of image features at point level. \VF{\,}involves projection of 3D voxels onto the image, followed by feature extraction within 2D ROIs and concatenation of pooled image features at the voxel level. In contrast to existing multimodal techniques, the proposed methods are single stage detectors which are simple and effective. Experimental results on the KITTI dataset demonstrate significant improvements over approaches using a single modality. Furthermore, our approach yields results competitive with the state-of-the-art multimodal algorithms on KITTI test set. In the future, we plan to train a multi-class detection network, and compare the current two-stage training with end-to-end training. 


%%%%%%%%%%%%%%%%%%%%%%%%%%%%%%%%%%%%%%%%%%%%%%%%%%%%%%%%%%%%%%%%%%%%%%%%%%%%%%%%
%\section*{ACKNOWLEDGEMENT}
%\label{sec:acknowledgement}
%We are grateful to our colleagues
%Russ Webb, Barry Theobald, and Jerremy Holland for their
%valuable input.

%%%%%%%%%%%%%%%%%%%%%%%%%%%%%%%%%%%%%%%%%%%%%%%%%%%%%%%%%%%%%%%%%%%%%%%%%%%%%%%%

%%%%%%%%%%%%%%%%%%%%%%%%%%%%%%%%%%%%%%%%%%%%%%%%%%%%%%%%%%%%%%%%%%%%%%%%%%%%%%%%
%\section*{APPENDIX}


%\section*{ACKNOWLEDGMENT}

%%%%%%%%%%%%%%%%%%%%%%%%%%%%%%%%%%%%%%%%%%%%%%%%%%%%%%%%%%%%%%%%%%%%%%%%%%%%%%%%

%References are important to the reader; therefore, each citation must be complete and correct. If at all possible, references should be commonly available publications.
{\small
	\bibliographystyle{ieee}
	\bibliography{mvxnet_arxiv}
}

\end{document}